# A Novel Image Dehazing and Assessment Method

Saad Bin Sami
*Department of Computer Science,*
*University of Karachi,*
Karachi, Pakistan
saadbinsami121@gmail.com

Abdul Muqeet
*Department of Computer Science,*
*Kyung Hee University,*
Seoul, South Korea
amuqeet@khu.ac.kr

Dr. Humera Tariq
*Department of Computer Science,*
*University of Karachi,*
Karachi, Pakistan
humera@uok.edu.pk

*Abstract—* **Images captured in hazy weather conditions often suffer from color contrast and color fidelity. This degradation is represented by transmission map which represents the amount of attenuation and airlight which represents the color of additive noise. In this paper, we have proposed a method to estimate the transmission map using haze levels instead of airlight color since there are some ambiguities in estimation of airlight. Qualitative and quantitative results of proposed method show competitiveness of the method given. In addition we have proposed two metrics which are based on statistics of natural outdoor images for assessment of haze removal algorithms.**

## I. INTRODUCTION

Outdoor images often suffer from haze, fog, and smog. When light travels through the medium, it strikes with the atmospheric particles which results in scattering and absorption of light and cause the irradiance to be attenuated along the line of sight of observer. Moreover, irradiance received by observer has additive noise too which causes the shift in scene colors and causes the atmospheric particles to behave like a source of light. Furthermore, it results in scene radiance to appear brighter. In case of dense haze, multiple scattering takes place which cause a slight burring effect too [1]. The combine effect of these phenomenon causes the captured image to appear low in color contrast and fidelity, which causes it to lose information. In addition, the performance and efficiency of different computer vision systems rely heavily on the quality of input image. Since different computer vision algorithms assumes that the input image is scene radiance [17]. Degraded image creates the hindrance in object detection, motion tracking, satellite imagery, aerial photography [9], autonomous driving and face recognition in CCTV security cameras [10]. This makes the task of haze removal to be inevitable before application of outdoor computer vision algorithms.

To solve these problems different methods have been developed for image dehazing. Tan et al. [8] observed that the haze-free image has a higher contrast than the hazy image. Thus, by optimizing a cost function in the Markov Random Field (MRF) they maximize the local contrast of the input hazy image but it causes the blocking artifacts around edges where scene depth changes abruptly.

Fattal in [2] proposed a method that derived transmission map based on the assumption that transmission and scene albedo are locally uncorrelated but it does not hold true for heavy haze. Moreover this method has a higher complexity and cannot be deploy on gray-scale images.

In 2009, He et al. in [17] proposed a novel method dark channel prior (DCP). This method utilizes the fact that with in the local patch of natural outdoor images there exist some pixels which have low intensities in at least one channel. However, the underlying assumption or statistics violates when there is a brighter object bigger than the patch size [3] or there is a bright object in the scene. But on the other hand, if we take the bigger patch size, the assumption that transmission is constant within a patch is violated at edges where scene depth changes abruptly.

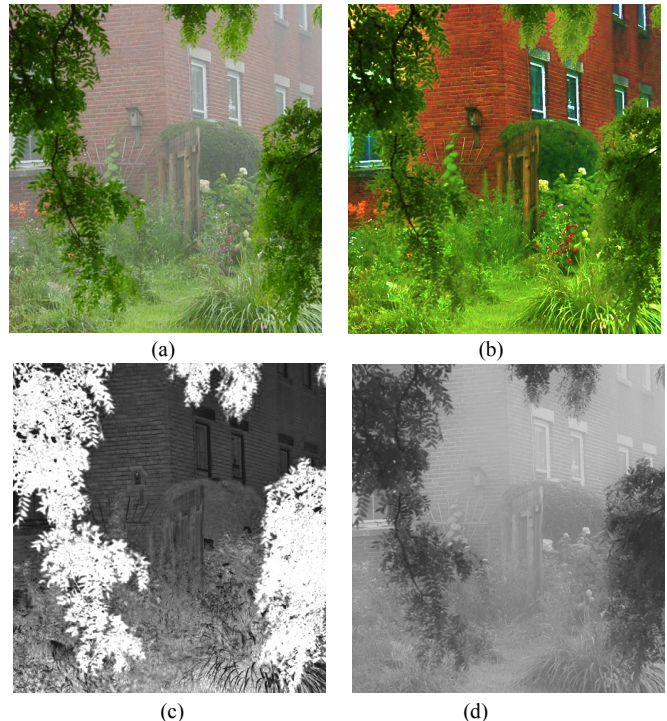

Figure 1. (a) Input hazy image. (b) Recovered haze free image. (c) Direct transmission map. (d) *K* map or haze map.

In [18] with in the local patch, color lines are fitted in RGB space, by searching for small patches with a constant transmission. Fattal recovered the dehazed image base on the fact that color lines in hazy images do not pass through the origin anymore, due to the additive haze component. In small patches that contain pixels that are equidistance from the camera, these lines are moved from the origin by the airlight at that distance. By fitting a line to each such local patch, the transmission within the local patch is evaluated using the distance the line has moved from the origin.

In 2016 Berman el at [4] proposed a non-local dehazing algorithm. Which is based on the observation that pixels of a haze free image forms tight clusters in RGB space. Pixels within the cluster are non-local i.e. they exist in different parts of image and have different scene depths and thus the transmission values. In the presence of haze these pixels translates its values to different transmission co-efficient values as they belong to different parts of image and thus have different thickness of haze.

In [5] dehazing method is proposed which is a learning based end-to-end system called DehazeNet which works on Convolutional Neural Network (CNN) based on deep architecture. It takes a hazy image as input, and outputs its transmission map. However, DehazeNet suffers from color distortion in poor illumination conditions [19].

In this paper we estimated transmission map and the unknown value $K$ associated with the additive component in the classical image degradation model presented in [1] and utilize the Guided Filtering. We call this constant the 'Airlight Co-efficient'. While many methods have been devoted to estimate the global airlight constant $A$, we on the other hand, estimate the value of $K$, see Fig. 1, which is the proportionality constant and its value depends on the exact nature and form of scattering function. The advantage of the method is that the outcome of this modification is the better color contrast as compared with other state of the art algorithms. Moreover, our method also works for dense haze too. Our approach is theoretically valid as the qualitative and quantitative evaluation of the results shows the robustness of the method given.

Rest of the paper is organized as follows. In section II, we will describe the image degradation model. In section III, we will discuss about the inaccuracies in estimation of airlight. In section IV and V, we will describe the dehazing algorithm and metrics for assessment of dehazing algorithms. Then, we will discuss qualitative and quantitative evaluation in section VI and VII respectively. And then we will conclude in section VIII.

## II. IMAGE DEGRADATION MODEL

### A. Model for Attenuation

In [1], it is noted that light reflected from the scene is scattered and absorbed by the atmospheric particles before reaching the camera. The amount of degradation of image depends on the size of particles in the medium and amount of scattering of light. Scattering of light increases with scene depth as given by the equation:

$$E(d, \lambda) = E(0, \lambda) \, e^{-\beta(\lambda) \, d} \qquad (1)$$

Where, $d$ is the scene depth $\beta$ is the total scattering co-efficient and it represents the ability of a medium to scatter a light of given wavelength $\lambda$ in all directions. $E(0, \lambda)$ represents the irradiance at $x = 0$ and $E(d, \lambda)$ represents the irradiance at observer i.e. $x = d$. It is the attenuated light that comes from scene to observer.

### B. Model for Airlight

According to [1] irradiance received by observer has additive noise too which causes the shift in scene colors. It often shifts the scene colors to gray-whitish and causes the atmospheric particles to behave like a source of light. It is given as:

$$L(d, \lambda) = K(1 - e^{-\beta(\lambda) d}) \qquad (2)$$

where, $K$ is the proportionality constant and its value depends on the exact nature and form of scattering function. It should be noted that due to unknown value of $K$ per pixel or point Narishman et al. [1] converted the above equation into:

$$L(d, \lambda) = L_{\infty}(1 - e^{-\beta(\lambda) d}) \qquad (3)$$

Where, $L_{\infty}$ is the light at horizon with distance $x = \infty$ and is often replace with symbol $A$. It is called global airlight constant. However, we deploy (2) in order to avoid inaccuracies in estimation of $A$, which are mentioned in section III.

### C. Combining Attenuation and Airlight Model

By combining (1) and (2) we get:

*Observed light = Attenuated light + Additive Airlight* (4)

If we let $e^{-\beta(\lambda) d} = t$ and $E(0, \lambda) = I$ and Observed light $= J$, then our model can be represented as:

$$I(x) = J(x) \, t(x) + K(1 - t(x)) \qquad (5)$$

Where, $x$ is the pixel co-ordinate, $t$ is the transmission coefficient. A higher value of $t$ represents that the given pixel is less attenuated and thus results in low additive noise and vice versa and $K$ is the airlight coefficient representing haze level.

$$L(d, \lambda) = K(1 - e^{-\beta(\lambda) d}) \qquad (2)$$

where, $K$ is the proportionality constant and its value depends on the exact nature and form of scattering function. It should be noted that due to unknown value of $K$ per pixel or point Narishman et al. [1] converted the above equation into:

$$L(d, \lambda) = L_{\infty}(1 - e^{-\beta(\lambda) d}) \qquad (3)$$

Where, $L_{\infty}$ is the light at horizon with distance $x = \infty$ and is often replace with symbol $A$. It is called global airlight constant.

### D. Combining Attenuation and Airlight Model

By combining (1) and (2) we get:

*Observed light = Attenuated light + Additive Airlight* (4)

If we let $e^{-\beta(\lambda) d} = t$ and $E(0, \lambda) = I$ and Observed light $= J$, then our model can be represented as:

$$I(x) = J(x) \, t(x) + K(1 - t(x)) \qquad (5)$$

Where, $x$ is the pixel co-ordinate, $t$ is the transmission coefficient. A higher value of $t$ represents that the given pixel is less attenuated and thus results in low additive noise and vice versa and $K$ is the airlight coefficient representing haze level.

## III. FLAWS IN AIRLIGHT ESTIMATION

Many different methods have been proposed to estimate $A$ using bright object or scene point in the scene. But does the white bright object always reflect the light color at horizon! Not always. This often results in overestimation of global airlight constant as mention in [20] and [14]. Tan in [8] took the maximum intensity value in the hazy image as the value of $A$, but clearly this would also overestimate $A$. In [16], airlight $A$ is estimated by taking the maximum of each channel i.e. $A \rightarrow [r^{max}, g^{max}, b^{max}]$, but this would also result in overestimation of airlight as it is highly probable that with in an image each channel would have at least one point where channel value is 1 so it is highly probable to get $A \rightarrow [1,1,1]$. Estimation of $A$ described in original DCP method is used in [21-27], [10] and many other methods. In this method, $A$ is estimated by first selecting the indices the top 0.1% pixels of dark channel and then using these indices the maximum intensity pixel in the hazy image is selected for

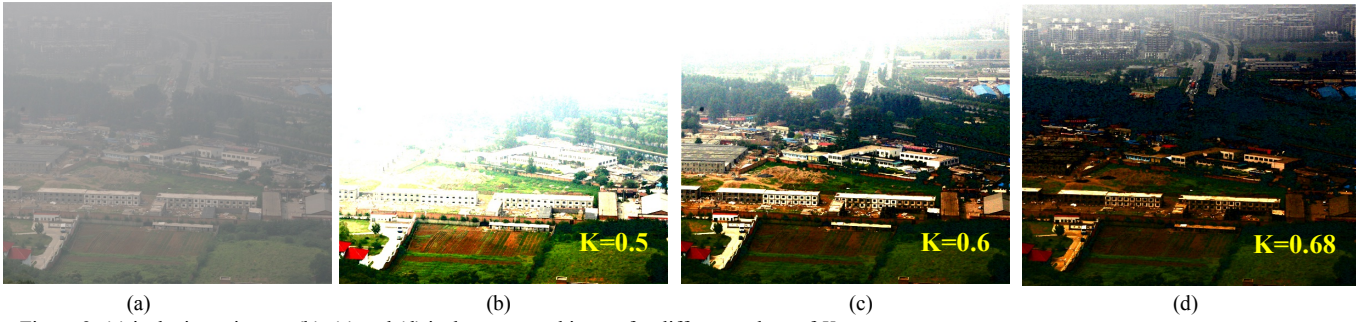

Figure 2. (a) is the input image. (b), (c) and (d) is the recovered image for different values of K.

A. But if the scene object is bigger than the size of patch then this would result in overestimation of A as mentioned in [20]. Moreover, if the accuracy of estimation of A heavily depends on size of the patch, bigger size of patch results in more accurate estimation. However, if we increase the patch size then computational cost will also be increased. Furthermore, patch size must be big enough to cover the bright object in the scene. So how big the patch size should be is still undecided because size of bright object does fluctuate from image to image.

In [15] top 0.2% pixels in the dark channel is selected for estimation of A. In [13] kim et al. estimate A based on the quad-tree subdivision. In this iterative method first the image is divided into four regions then the region with maximum score will be selected for further decomposition until the size of the selected region is less than the threshold value. Where the score of each region is calculated by subtracting average pixel values from the standard deviation of the pixel values with in that rectangle. But this method also fails when there is a bright object in the scene as mentioned in [14]. Moreover, there are a lot of different algorithms that have been used to segment sky and foreground and then selecting pixels from sky as an estimate to A. This seems to satisfy the condition that airlight should be a light coming from infinity or maximum depth, given in [1]. However, we often encounter images without sky region, in that case we have to select a pixel from foreground as an estimate to A which could be prone to error. We, on the other hand, estimate the value of airlight co-efficient K to avoid any of these ambiguities.

## IV. HAZE REMOVAL

### A. Estimation of Transmission map

We estimate the transmission map based on the statistics that of natural outdoor images that the product of minimum channel of haze free image and its transmission map of hazy image will approach to zero so it is negligible. First we estimate the transmission map as:

From (5),

$$I(x) = J(x)\,t(x) + K(x)\,(1 - t(x))$$

Taking minimum of each channel we get,

$$\min_{c \in \{r,g,b\}} (I^c(x)) = \min_{c \in \{r,g,b\}} (J^c(x))\,t(x) + K(x)\,(1 - t(x))$$

Here the term $\min_{c \in \{r,g,b\}} (J^c(x))\,t(x)$ can be neglect because in case of haze the transmission map have low values and the minimum operation applied on each color vector would also result in low values. In addition, the product of these values also lower the term. In the next section we will discuss the

quantitative validation of this preposition. So the above equation can be reduced to:

$$\min_{c \in \{r,g,b\}} (I^c(x)) = K(x)\,(1 - t(x))$$

Solving above equation for $t(x)$ would result in

$$\widetilde{t}(x) = 1 - \omega(\min_{c \in \{r,g,b\}} (\frac{I^c(y)}{K})) \qquad (6)$$

Where, $\omega$ is set to be 0.95 to permit some haze in order to preserve natural look. Since above formula is derived by neglecting the lower value term $\min_{c \in \{r,g,b\}} (J^c(x))\,t(x)$, this would roughen the transmission estimation. Therefore, normalizing it with a smaller value will compensate for this. So we use:

$$\widetilde{t}(x) = \frac{1 - \omega(\min_{c \in \{r,g,b\}} (\frac{I^c(y)}{K}))}{1 - \beta} \qquad (7)$$

Where, $\beta$ is defined as:

$$\beta = \underset{c \in \{r,g,b\}}{mean}(I^c) - \min_{c \in \{r,g,b\}} (I^c) \qquad (8)$$

In (7), the only unknown variable is K. To estimate the value of K, we did some experiments for different values of K and the results are shown in Fig. 2. From these results it is clear that the value of K is different for different regions. To put it in context we can say that the value of K changes with the value of haze, as the haze or fog increases the value of K also increases and vice versa or it changes with scene depth and for homogenous region where haze level remains same the value of K also remains same. So we can say that the value of K represents the level of haze. And the value of K is local to the region and not global that is why in Fig. 2(b) the foreground region is recovered well and the value $K = 0.5$ suited well for this region and in background region this value is underestimated and hence result in burnt (bright) pixels. Similarly, in case of Fig. 2(d) the background region is recovered better as compare to foreground and the value of $K = 0.68$ is overestimated for foreground region. Mathematically we can say that:

$$K = f(haze\ (depth))$$ (9)

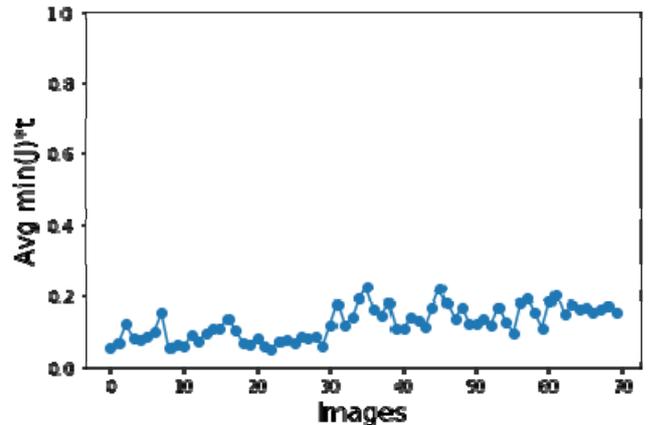

Figure 3. 70 outdoor hazy and haze free images were taken from Nitri Dehazing competition 2018[11] and 2019[12]. Then we estimate the transmission map by using Dark Channel Prior[17] and then calculate the product of transmission map and minimum of J for each image set (I,J). Then the average of each of the product was plotted which shows that the product has low values so we can neglected this product.

### B. Estimating Airlight Co-efficient K

From above discussion it is clear that haze level remains locally with in a region and changes with scene depth. So we need a homogenous value of $K$ locally. For this we exploit the property of natural outdoor images that grayscale image of the hazy image will have higher values as compare to its corresponding non-hazy image. Therefore, the estimation of $K$ is as follows. First we estimate the grayscale of the given hazy image.

$$C(x) = \frac{I^r + I^g + I^b}{3} + \alpha$$ (10)

Where, $(r, g, b)$ is the red, blue and green color channel of the input image $I$ and $\alpha$ is the constant used to avoid color distortion whose value is estimated as:

$$\alpha = \mu_I - \mu_{Mc}$$ (11)

Where, $\mu_I$ is the mean of the input image $I$. It is the single value and $\mu_{Mc}$ is the mean of minimum channel of $I$. Mathematically,

---

**Algorithm 1.1: Computing of Haze Removal Metrics**
**Input: $I(x)$, $J(x)$**
**Output: $\beta$**
1: $I_A(x) = I(x) - A$
2: Convert $I_A(x)$ to spherical coordinates to obtain the required $[r(x), \emptyset(x), \theta(x)]$.
3: Cluster the pixel according to $[\emptyset(x), \theta(x)]$. Each cluster $i$ is a haze line.
4: For each haze line $i$ do:
    5: Select indexes of cluster $i$ from $I(x)$ and using these indexes select the indexes of $J(x)$. Using these indices compute magnitude of each color vector of $I(x)$ and $J(x)$ in RGB space to get the corresponding magnitude set $(m_{I1}, m_{I2}, m_{I3}, \ldots\ldots, m_{In})_i$ and $(m_{J1}, m_{J2}, m_{J3}, \ldots\ldots, m_{Jn})_i$ respectively. Where $n$ is the number of pixels in a cluster and $i$ is the cluster index and $I$ and $J$ represents the pixels of hazy and haze free images.
    6: Now compute the standard deviation of each of the magnitude set of $I(x)$ and $J(x)$ to get the cluster deviation set for $I(x)$ and $J(x)$ as $CD_i^I$ and $CD_i^J$ where $i$ is the number of clusters.
    7: Calculate the Mean Square Error for both $CD_i^I$ and $CD_i^J$ as:

$$\beta = \frac{\sum_{y=0}^{i} \left(CD_y^I - CD_y^J\right)^2}{y}$$ (12)

---

$$\mu_I = mean\ (\underset{c \in \{r,g,b\}}{mean}(I^c))$$ (13)

Where, $\underset{c \in \{r,g,b\}}{mean}(I^c)$ returns the mean of each channel and

$$\mu_{Mc} = mean\ (\underset{c \in \{r,g,b\}}{\min}(I^c))$$ (14)

As mentioned above that haze level remains same locally so we take the average filter to smoothen $C(x)$.

$$C(x)_{avg} = \underset{y \in \Omega(x)}{average}(C(y))$$ (15)

Where, $\Omega(x)$ is the filtering window centered at pixel $x$. Now we apply guided filter [7] for edge preserving properties.

$$K(x) = guidedfilter\ (I, p = C(x)_{avg})$$ (16)

Finally we set the lower bound on $K$ to avoid it being too low.

$$K(x) = \max(K(x), K_0)$$ (17)

Where, $K_0$ is set to be 0.75. Although its value varies from 0.75 to 0.87, for suitable results.

### C. Validating Approximation of Transmission Estimation

In section VI-A, we neglect the term $\underset{c \in \{r,g,b\}}{\min}(J^c(x))\,t(x)$ while deriving the formula of transmission map. In this section we will validate this approximation quantitatively. To prove this we need the haze free image of the corresponding hazy image. So we take the 70 outdoor haze free and corresponding hazy image provided in the [11] and [12] and validate our results. We estimate the transmission map of the hazy image by using the [17] and [7] and take the product of minimum channel of haze free image and the transmission map of the corresponding hazy image for each of 70 image set. Fig. 3 shows the plotting of average of the product of $J$ and transmission map for each image set $(I, J)$ which indicates that this product has low values globally. So we can neglect this product as we have. And the mean of the 70 terms of $\underset{c \in \{r,g,b\}}{\min}(J^c(x))\,t(x)$ is found to be **0.12**, which seems to be a small value. But this approximation will result in slight color distortion.

## V. PROPOSD METRICS

To test the performance of any image processing application it is often desired to have both noisy image and the ground truth image. This type of image assessment techniques are called full-reference image quality assessment and the classical approaches are mean square error (MSE), peak signal to noise ratio (PSNR) and structural similarity index (SSIM). However, it is impossible to get the ground truth image prior to haze removal in real-time applications. This makes the assessment of haze removal algorithms to be no-reference quality assessment where we do not have a ground truth image prior to haze removal. The famous no-reference quality assessment methods presented in [6] are:

1) Rate of newly visible edges $e$ between input of output images.

2) The ratio of visible edge's gradient $\bar{r}$ between input and output images.

3) The percentage of pixels $\sigma$ that become completely black or white after haze removal.

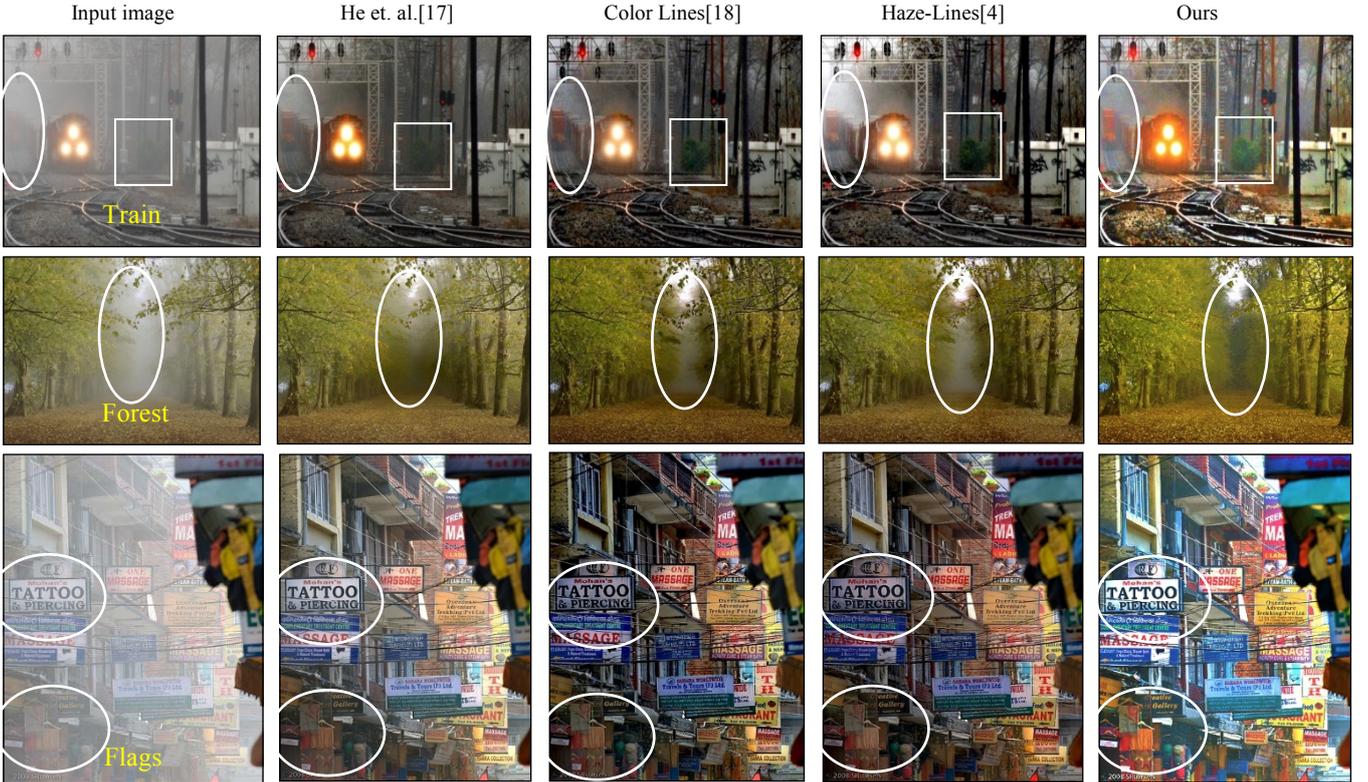

Figure 4. Qualitative comparison of proposed haze removal results with other state of the art methods. First column is the input hazy image. second third , fourth and fifth columns are the results of He et.. al.[17], Color-Lines[18], Haze-Lines[4] and ours respectively.

These metrics gives a good indication of how much image contrast is recovered after haze removal, In addition these metrics are not based on statistics of natural outdoor images. And it is always desired to find a way to assess haze removal algorithm based on full-reference image quality assessment method. Furthermore, these methods do not quantify color distortion after haze removal. To solve these problems, in this paper, we have proposed metrics based on [4] and [17]. The proposed metrics works similar as a full-reference image quality index and rely on the statistics or properties of natural outdoor haze free images and haze theory. The proposed metrics are based on two priors:

### A. Based on Theory of Dark Channel Prior

In this paper, we proposed a metrics based on [17]. This metrics is based on the fact that dark channel gives clue to haze. So the recovered image must have a dark channel with low intensity. Hence the difference in the dark channel of image before haze removal and after haze removal will give the clue about how much haze is removed. So we take the mean square error of the dark channels before and after haze removal. In foreground region after haze removal dark channel have low intensities and in background or sky region dark channel would have high values after haze removal. Mathematically we can express it as:

$$\alpha = \frac{\sum_{x=0}^{M-1} \sum_{y=0}^{N-1} \left( D(x,y) - \hat{D}(x,y) \right)^2}{MN} \quad (18)$$

Where, $M$ and $N$ are dimensions of dark channel. $D$ is the dark channel before haze removal and $\hat{D}$ is the dark channel after haze removal. This metrics gives good indication of haze removal but it could mislead when there is color distortion. For this we have proposed a metric based on [4].

### B. Based on Theory of Haze Line

It was noted in [4] that pixel value of haze free natural images forms tight clusters in RGB space. Pixels in a single cluster are often dispersed spatially in the image. And when haze obstructs the camera, these pixels translate to different coefficients based on the haziness in that region. This causes these tight clusters to translate to lines (called haze lines). So the difference of standard deviation of magnitude of color vector in each haze line and haze free color cluster will give the clue to how much haze is removed, since these haze lines should form tight clusters again after haze removal. See Table I, for quantitative evaluation of these metrics with some images. The algorithm for evaluation of this metrics is given in algorithm **1.1**.

TABLE I. QUANTITATIVE EVALUATION

| Images | Assessment | [17] | [4] | [18] | ours |
|---|---|---|---|---|---|
| Train | $e$ | 1.2861 | 1.1717 | **1.3383** | 1.324 |
| | $\bar{r}$ | 4.0201 | **5.6386** | 4.8805 | 0.01542 |
| | $\sigma$ | 0.05931 | **7.4657** | 5.0149 | 7.0858 |
| | $\alpha$ | 0.07711 | 0.12554 | **0.13722** | 0.07329 |
| | $\beta$ | 0.99473 | 0.97048 | 1.0863 | **1.17609** |
| Flags | $e$ | 0.22004 | 0.21343 | 0.13943 | **0.27224** |
| | $\bar{r}$ | 5.7183 | **6.3301** | 5.7164 | 0.34137 |
| | $\sigma$ | 0.1745 | **7.3968** | 4.102 | 7.0837 |
| | $\alpha$ | 0.14494 | **0.23783** | 0.23586 | 0.16563 |
| | $\beta$ | 0.22450 | 0.06229 | 0.05363 | **0.24975** |
| Forest | $e$ | 0.29471 | **0.42131** | 0.39373 | 0.28866 |
| | $\bar{r}$ | 4.1309 | **4.4598** | 4.4351 | 2.1559 |
| | $\sigma$ | 0.14409 | 3.4485 | 2.0565 | **5.4797** |
| | $\alpha$ | 0.04339 | **0.06799** | 0.06375 | 0.05174 |
| | $\beta$ | 0.45715 | **0.65539** | 0.60523 | 0.47846 |

## VI. QUANTITATIVE EVALUATION

The quantitative evaluation is presented in Table I. We have compared our proposed method with other state of the art method by selecting most famous images used in haze removal algorithms for comparison. Results show that proposed methodology has performed better than [17]. However we found in some cases haze has intact after haze removal due to some inaccuracy in estimation of K-map. Although we have proposed a method for empirical estimation of airlight co-efficient $K$ we believe that more theoretical grounds are needed for future development.

## VII. QUALITATIVE EVALUATION

The qualitative comparison of proposed method is presented in Fig. 4. Important regions are highlighted with white rectangles for aid in comparison. In all haze removal results we have fixed the patch size to be 9 and $k_0 = 0.8$. While comparing we can say that there are cases when our algorithm has worked better than other methods. In image train the overall haze removal, brightness and visibility is far better than other state of the art algorithms. Especially if we observe a bush near train and containers on the left side of train. In forest image our method has removed haze much better than other algorithms as it is highlighted. If we view the picture on large scale, we find that it suffers from color distortion but lesser than [4] and [18] as highlighted. Although color distortion is supposed to be a problem in haze recovery, in cases with dull light this color exaggeration also helps in identifying colors in a better way as shown in the foreground region of flag image.

## VIII. CONCLUSION

On concluding, we can say that the observation about airlight map brought in this research shows that the accurate estimation of airlight coefficient map will lead to best haze removal results and color contrast. The estimation of airlight co-efficient method proposed in this paper shows some good results which is evident from quantitative and qualitative analysis but we also found that in there are some cases where this method underperformed state of the art methods. And this creates a gap for other research community to find the better estimations for airlight co-efficient. One plus point of our results is that they has good visibility and brightness. The other advantage of this research is the metrics which is based on the haze theory i.e. it quantifies how much haze is removed, which should be the metrics for assessment of haze removal algorithms.

## IX. REFRENCES